# Modeling of Time-varying Wireless Communication Channel with Fading and Shadowing

Youngmin Lee, Xiaomin Ma, and Andrew S.I.D. Lang
College of Science and Engineering, Oral Roberts University, Tulsa, OK 74171, USA
Emails: ttmn_ym@oru.edu, xma@oru.edu, and alang@oru.edu

*Abstract*—The real-time quantification of the effect of a wireless channel on the transmitting signal is crucial for the analysis and the intelligent design of wireless communication systems for various services. Recent mechanisms to model channel characteristics independent of coding, modulation, signal processing, etc., using deep learning neural networks are promising solutions. However, the current approaches are neither statistically accurate nor able to adapt to the changing environment. In this paper, we propose a new approach that combines a deep learning neural network with a mixture density network model to derive the conditional probability density function (PDF) of receiving power given a communication distance in general wireless communication systems. Furthermore, a deep transfer learning scheme is designed and implemented to allow the channel model to dynamically adapt to changes in communication environments. Extensive experiments on *Nakagami* fading channel model and *Log-normal* shadowing channel model with path loss and noise show that the new approach is more statistically accurate, faster, and more robust than the previous deep learning-based channel models.

## I. INTRODUCTION

In wireless communication systems, a signal sent from one node to another may experience significant degradation or attenuation due to fading, shadowing, path loss, or interference from other transmitting nodes [1]. Thus, building a channel model is crucial for analyzing how the channel affects the transmitted signal. This analysis is essential in designing communication parameters and optimizing network configurations in real time. [2]-[4]. Most practical channels are complicated and hard to evaluate theoretically with closed-form formulas. Recently, several pure data-driven approaches to channel modeling using deep learning neural networks (DLNNs) have been proposed to approximate the channel output (*y*) distribution given channel input *x*: $p(y|x)$ using Generative Adversarial Networks (GANs) excluding the effect of coding/decoding, modulation/demodulation, and signal preprocessing [5]-[8]. Since the DLNN models only consider the fading effect without accounting for the impact of the path loss effect on the transmitting signal in the channel, the application scope of this model is somewhat limited. These models have been applied to the classification of modulated signals in relatively simple channels. Most recently, a channel modeling mechanism using two deep learning models, GANs and feedforward neural networks (FNN), was proposed and verified to approximate stochastic characteristics of a more general channel model with fading [9]. New metrics were designed in the paper to evaluate the precision and effectiveness of the channel model. However, the objective (or loss) function used in the training of the neural networks, such as a sum-of-squares or cross-entropy error function, could lead to network outputs that approximate the conditional average of the target data, conditioned on the input vector [10]. For regression problems involving the prediction of continuous variables, the conditional averages provide only a very limited description of the properties of the target data. Therefore, the accuracy and robustness of such deep learning-based channel models are not assured.

In this paper, a new way of channel modeling that combines deep learning neural networks and the mixture density model (DMDN) [10] is proposed and tested to approximate stochastic characteristics of a more general channel model in real time. Compared with the existing methods for channel modeling, the main contributions of this paper are: 1) The proposed channel model approximates an arbitrary probability density as a linear summation of multiple Gaussian kernel functions with distinctive means and variances. Then, the deep-learning neural networks are configured to learn the coefficients of the kernel functions. 2) Comparisons of the DMDN model with the previous channel modeling using cGAN and FNN for modeling of *Nakagami* fading channel and *Log-normal* with path loss are conducted, and it is found to have significantly higher accuracy and faster convergence speed. 3) A deep transfer learning scheme is designed and implemented to allow the channel model to dynamically adapt to changes in communication environments. The paper is organized as follows. Section II briefly reviews wireless communication systems, channels, and channel modeling. Section III describes the principle, structure, and implementation of DMDN channel modeling. Section IV outlines the experimental setup used to validate the accuracy and effectiveness of the DMDN modeling scheme, followed by a presentation of the numerical results and discussions. Conclusions are presented in Section V.

## II. SYSTEM DESCRIPTION AND PROBLEM FORMULATION

Many wireless communication systems have been developed to enable seamless communication across diverse platforms with high quality of services (QoS). The communication systems include but are not limited to, upgraded Wi-Fi networks for high-speed internet access, ad hoc networks for mission-critical applications, and 5G/6G cellular networks for personal communications. These systems leverage innovative technologies such as advanced coding methods in the physical layer (e.g., low-density parity checking (LDPC)), high-speed modulation and coding schemes (MCS) (e.g., quadrature amplitude modulation (QAM) 1024), Multi-user Multiple-input Multiple-output (MU-MIMO) and orthogonal frequency division multiplexing access (OFDMA), etc.

The QoS of a communication system is determined by the random wireless channel that alters the signal from the transmitter to the destination. Wireless communication channels are characterized by various impairments, including fading, shadowing, path loss, noise, and interference from other nodes' transmissions. Fading refers to the rapid variations in signal strength due to factors like multipath propagation. Shadowing occurs due to obstacles in the signal path, leading to signal attenuation in specific areas. Path loss represents the gradual reduction in signal strength over distance, influenced by factors such as free-space loss and absorption by atmospheric gases. Noise and interference occur when nodes receive multiple signals from different transmitters or other sources in the same channel. Understanding and mitigating these challenges is crucial for the analysis and design of robust wireless systems.

Given a transmitting signal power strength $P_{tx}$ and a communication distance $d$ between a sender and a receiver, the immediate receiving power $P_{rx}$ is a random function of $P_{tx}$ and $d$. Then, channel fading/shadowing is characterized by the cumulative distribution function (CDF) and the probability density function (PDF) of power or strength of the signal at a receiver with distance $d$ away from a source node:

$$F_{P_{rx}|d}(y,t) = \Pr(P_{rx} \leq y|d,t), \quad (1)$$

$$f_{P_{rx}|d}(y,t) = \frac{dF_{P_{rx}|d}(y,t)}{dy}, \quad (2)$$

and

$$\overline{P_r(d)} = PL(P_t, d, t) + N_G, \quad (3)$$

where $\overline{P_r(d)} = \mu_i(d)$ is average received signal strength or power as a path loss function $PL(P_t, d, t)$ of transmitting power and communication distance $d$, and $N_G$ is added noise. Notice that CDF and PDF are functions of communication distance and depend on observation time $t$. Also, note that $f_{P_{rx}|d}(y,t)$ is independent of communication modulation modes and coding schemes. The impact of the modulation/demodulation, coding/decoding, and interference from other nodes on the transmission errors can be evaluated separately in the form of frame (bit) error rates as a function of Signal-to-Interference & Noise Ratio (SINR).

The objective of the modeling is to collect or generate data pairs of $(x_i, y_i) = (d, P_r)$ without considering the modulation, coding, or signal processing type and train the DLNNs to approximate channel statistical impacts on the transmitting signals. The measurement of the signal pairs is conducted on each node distributively in the network and reported to a local edge platform that covers the network. The data collection on each node proceeds as follows:

a) Decide ranges of modulated signal parameters $P_t$;
b) Given two nodes with distance value $d_i$, apply the input signal $P_t$ to the channel to be evaluated;
c) Measure $P_r$ and distance $d_i$;

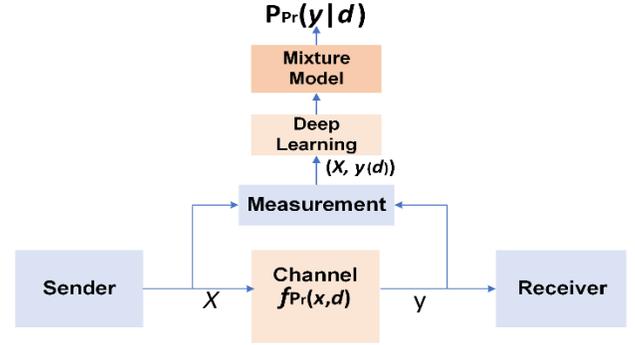

Fig. 1 Structure of DMDN for channel modeling

d) If no interference from other nodes is detected, send data pairs to the local edge platform; otherwise, skip them.
e) Repeat (a)-(d) until the number of data pairs is sufficient;
f) Find other node pairs, repeat (a)-(e).

## III. DMDN CHANNEL MODELING

### A. Principles of the DMDN Approach

We propose a new approach to adaptive estimation of changing fading/shadowing in the wireless channels from data collected from the measurements, which combines a deep learning neural network with a mixture density model from [10]. As shown in Fig. 1, instead of training a deep learning network using data measured and collected directly from the wireless channel for evaluation of the probability distribution [9], we adopt and modify the mixture density model [10] to approximate an arbitrary probability density of an observed channel as a function of receiving power and communication distance. In our deep learning mixture density network (DMDN) model, the PDF of the receiving power is represented as a linear combination of Gaussian kernel functions in the form

$$f_{P_{rx}|d}(x) = \sum_{i=1}^{m_c} (a_i(d)) \phi_i(P_{rx}|d), \quad (4)$$

where $m_c$ is the number of components in the mixture, $\alpha_i$ are mixing coefficients, and the function $\Phi_i$ represents the conditional density of the receiving power strength $P_{rx}$ for the $i$th kernel, which is expressed as

$$\phi_i(P_{rx}|d) = \frac{1}{(2\pi)^{\frac{1}{2}} \sigma_i(d)} exp\left\{-\frac{\|P_{rx} - \mu_i(d)\|^2}{2\sigma_i(d)^2}\right\}, \quad (5)$$

where $\mu_i(d)$ and $\sigma_i(d)$ represent the center and variance of the $i^{th}$ kernel. McLachlan and Basford have demonstrated that a Gaussian mixture model with kernels given above can approximate any given density function to arbitrary accuracy, provided the mixing coefficients and the Gaussian parameters are correctly chosen [10]. Therefore, the mixture density model can be used to characterize any wireless channel with a general density function. The deep learning neural network is built and trained by collected data pairs to determine the parameters in the mixture model, denoted as $z = \{z_i^\alpha, z_i^\sigma, z_i^\mu\}$. The training of the deep learning neural network can be achieved by maximizing the likelihood that the model gives rise to a particular set of data points. The error function is defined by taking the negative logarithm of likelihood.

$$E = \sum_q -ln\{\sum_{i=1}^{m_c} \alpha_i(d^q)\phi_i(P_{rx}^q|d^q)\} \quad (6)$$

Fixing $P_{tx}$ strength, each receiver can measure the power strength of the receiving signal and the distance between the receiver and the sender. The data pairs are used to train a deep-learning neural network to catch up with the stochastic properties of the channel in the form of $f_{P_{rx}|d}(x)$. An advanced version of deep learning networks will be selected and tested for estimation of parameters in Eq. (5). To learn the probability distribution of the channel fading/shadowing that could change in real-time, a transfer learning mechanism will be applied to the deep learning structure: train core deep learning structure, take the structure and learned weights as initialization, then train all layers with new measured data to adapt the dynamic changing of the wireless channel.

### B. Implementation of DMDN for Channel Modeling

The pseudo-code of the DMDN for channel modeling is shown below. Note that we used the same parameters for transfer learning. DMDN with the given parameters takes 9-11 seconds per epoch under Google Colab CPU (Intel® Xeon® CPU @ 2.20GHz.). To avoid invalid loss during the training, a custom Keras Callback function that monitors the loss values and reinitiates the training process is called if any NAN loss gets detected.

---

**Algorithm 2** Build Model - DMDN

---

**Input:** genuine data $x_i$, distance $d_i$
**Output:** generated data $MDN(d_i)$
  1: **BUILD** *DMDN* by following
  2:   x = Input($d_i$)
  3:   **FOR** num_layer
  4:     x = Dense(num_unit, activation=Relu)(x)
  5:   **END FOR**
  6:   x = Dense(num_MDN_param, actovation=None)(x)
  7:   x = MixtureNormal(num_component, event_shape)(x)
  8: **COMPILE** MDN with
       Optimizer=Adam, Learning Rate=lr, Batch Size=bats,
       Loss= -MDN.log_prob($y_{true}$)

num_layer = 8, num_unit = 256, lr = 0.005, bats = 10,000
num_components = 8, event_shape = [1],
num_MDN_param=num_layer+2×num_component×event_shape

---

### C. Setup of Experiments for Testing and Comparisons

To test the effectiveness of the proposed DMDN channel modeling method and compare it with other deep-learning channel models, we chose two typical theoretical models as benchmarks: the *Nakagami* fading channel model with exponential path loss and the *Log-Normal* shadowing model with path loss. Data generated from the two distributions are utilized to train the DMDN.

Specifically, for the *Nakagami* model, PDF of strength $P_{rx}$ conditioned on communication distance $d$

$$f_{P_{rx}|d}() = \frac{1}{\Gamma(m)} 2\left(\frac{m}{\overline{P_r(d)}}\right)^m x^{2m-1} exp\left(-\frac{mx^2}{\overline{P_r(d)}}\right) \quad (7)$$

where $\Gamma(m)$ is the Gamma function, and $m$ is the fading parameter, and

$$\overline{P_r(d)} = PL(P_t, d) = P_t\eta\left(\frac{d_0}{d}\right)^\alpha \quad (8)$$

The normal Gaussian distribution function is adopted to generate the channel noise $N_G$ with the mean of the distribution $N_{G_{mean}}$ and the variance $N_{G_{var}}$. In the experiment with *Nakagami*, $\eta$, $N_{G_{mean}}$, and $N_{G_{var}}$ are scaled with $10^{14}$ for the purpose of convenience.

A *Log-normal* shadowing channel model is formulated by the following equations:

$$f_{P_{rx}|d}(x) = \begin{cases} \frac{1}{\delta_1\sqrt{2\pi}} exp\left(-\left(\frac{P_{rdB}(d)-x}{\delta_1\sqrt{2}}\right)^2\right), & d_0 \le d \le d_c \\ \frac{1}{\delta_2\sqrt{2\pi}} exp\left(-\left(\frac{P_{rdB}(d)-x}{\delta_1\sqrt{2}}\right)^2\right), & d > d_c \end{cases} \quad (9)$$

$$P_{rdB}(d) = \begin{cases} P(d_0) - 10\alpha_1 log\left(\frac{d}{d_0}\right) + X_{\delta 1}, & d_0 \le d \le d_c \\ P(d_0) - 10\alpha_1 log\left(\frac{d_c}{d_0}\right) - 10\alpha_2 log\left(\frac{d}{d_c}\right) + X_{\delta 2}, d > d_c \end{cases} \quad (10)$$

where $P(d_0)$ is the signal power strength as calculated using the free space model with path-loss at the reference distance $d_0$, which is expressed as

$$P(d_0) = 10log\left(P_t \frac{G_t G_r \lambda^2}{(4\pi d_0)^2}\right) = 10log(P_t\eta/d_0^2),$$

$d_c$ denotes the critical distance for a given road type. $\alpha_1$ and $\alpha_2$ are the path-loss exponents at two reference distances, respectively. $X_{\delta 1}$ and $X_{\delta 2}$ describe the random shadowing effects, which is a Gaussian distribution with mean 0 and variance $\delta_1^2$ and $\delta_2^2$, respectively.

We applied log scaling (Eq. (10)) for $P_r(d)$ before feeding them to the DLNNs for the following reasons: 1) The distribution of $P_r(d)$ is positively skewed; 2) and its range varies dramatically with $d$. Then, the values were scaled back after the DLNN models.

$$scaled\ x = \log_{10}(x + coef) \quad (11)$$

Since log scaling requires positive input, we set the coefficient 2 for *Nakagami* and 435 for *Log-Normal* to prevent any negative values within the d range.

For evaluation and comparison, the following metrics are adopted from [9] or newly defined to objectively evaluate and compare the modeling performance.

$GeneratedMean(P_r(d))$, and $GeneratedVar(P_r(d))$ where $n$ is the number of $P_r(d)$ for each category ($P_{t_{array}}$ or $d_{array}$).

$$\bar{x} = \Sigma x/n \quad (12)$$
$$s^2 = \Sigma(X - \bar{x})^2/(n - 1) \quad (13)$$

*Overlapped Area* (*OA*), which exhibits how the PDF of receiving signal from the DLNN model synchronizes with the ideal PDF (with noise). The process to determine the overlapped area is: 1) Generate *genuine data* from Eq. (14) (apply the noise if needed) and *generate data* from the DMDN by valid categories. 2) Calculate the Gaussian Kernel Density Estimation for both *genuine* and *generated data* for each category.

$$OA_{local} = \sum_{i=1}^{k-1} \frac{1}{2}(f_{lower}(x_i) + f_{lower}(x_{i+1}))\Delta x, \quad (14)$$

and

$$f_{lower}(x_i) = Min\left(f_{\sigma\sim genuine}(x_i), f_{\sigma\sim generated}(x_i)\right), (15)$$

$$f_\sigma(x) = \frac{1}{n\sigma}\sum_{i=1}^n K(\frac{x-x_i}{\sigma}), \quad (16)$$

where $K$ is the Gaussian kernel and $\sigma = 0.3$ in this experiment. $\{x_i\}_{i=1}^{i=k}$ is a partition or $[s_{min}, s_{max}]$ with $x_1 =$

$s_{min}$ and $x_k = s_{max}$ and $k$ is the number of $P_r(d)$ for each category ($d$). We use $k$ as $\frac{n(P_r(d))}{10}$ in this research since we verified that this makes no critical difference ($\pm 0.01$ with *OA*) under large enough data. Here, we adopt a new metric: *Modified Overlapped Area* (*MOA*) that can be applied to the evaluation of any type of channel distribution:

$$MOA = Average\ OA - 2 \times \sigma, \quad (17)$$

where $\sigma = \sqrt{s^2}$, where $s^2$ is from Eq. (13). Note that the coefficient of $\sigma$ can be changed by circumstance.

By considering the standard deviation of *OA* for each d with the *Average OA*, *MOA* measures the correspondence of PDF between genuine and generated data and the model's consistency throughout $d$.

*Scaled Percent Error* (*ScaledPE*) indicates the Average Scaled Percent Error if it does not specify its categorical value, which considers the variance more significant [9]:

$$ScaledPE = (PEMeanAvg * 0.3 + PEVarAvg * 0.7) / 2, \quad (18)$$

where *PEMean*, and *PEVar* are the percent error for the mean and variance between the ideal and generated data:

$$PE = \left|\frac{Ideal - Generated}{Ideal}\right| \times 100. \quad (19)$$

## IV. NUMERICAL RESULTS AND DISCUSSIONS

### A. DMDN with Nakagami Fading Channel Model

We apply DMDN to the modeling of the *Nakagami* 1 (*N1*) fading channel with the following configuration: while $P_t$ is fixed to $0.28183815W$, $d$ has 30 values ($d = \{10, 20, 30, \dots, 300\}$), and $m$ has two possible values ($m = 2$ if $d \leq 140$ else $m = 1$). The rest of the variables are $\eta = 7.29 \times 10^{-14}, d_0 = 100, \alpha = 2$. Then the Gaussian noise is applied to the channel while $N_{G_{mean}} = 1.256 \times 10^{-15}W$, and $N_{G_{var}} = 1 \times 10^{-15}$ and $1 \times 10^{-16}$. The numerical results in terms of *OA* and *ScaledPE* are compared with cGANs and FNNs (MSE and RMSE loss) in [9] while keeping the same amount of train, validation, and test data size of 0.2, 0.16, and 0.1 million, respectively. The DMDN is trained with 15 epochs for each data set, and the same process is iterated 10 times. One of the issues of DMDN is that the model could yield NAN loss value easily, especially when the DMDN's depth is not enough compared to the complexity of the input data. Therefore, we implement a custom Keras Callback function that monitors the loss values and reinitiates the training process if any NAN loss gets detected.

Fig. 2 and Fig. 3 show the *OAs* and *ScaledPE* of DMDN and their comparisons with that of FNN and cGAN, which are obtained in [9]. According to the experiment implementing the DMDN model on the *Nakagami* channel, *OA* values of MDN with all noise variance cases outperformed FNN and cGAN, which indicates the effectiveness and superiority of DMDN. However, DMDN has higher *ScaledPE* values than the other models in most cases (See Fig 3; note that the lower *ScaledPE* indicates better performance in that perspective). This observation is because *ScaledPE* and *MOA* measure the model's performance from different perspectives. Although *ScaledPE* considers the model's consistency by doing average, it majorly focuses on determining how close the mean and variance of the distributions of the data by $d$. *MOA* evaluates the overall agreement between the PDF of genuine and generated data and its consistency through $d$. Still, it does not consider the mean and variance of the ideal data distributions. Our previous research [9] indicates that the performance of both FNN and cGAN decreases as the ratio of noise to signal distributions and $d$ increases. However, DMDN can hold strong performance consistency over all distances and noise levels, as shown in Fig. 2. Note that the best DMDN are selected by *MOA*, and the FNN and cGAN are by *ScaledPE*.

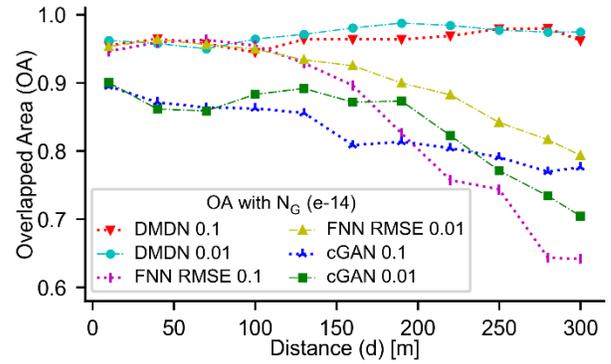

Fig. 2 *OA* of DMDN, FNN, and cGAN as functions of communication distance.

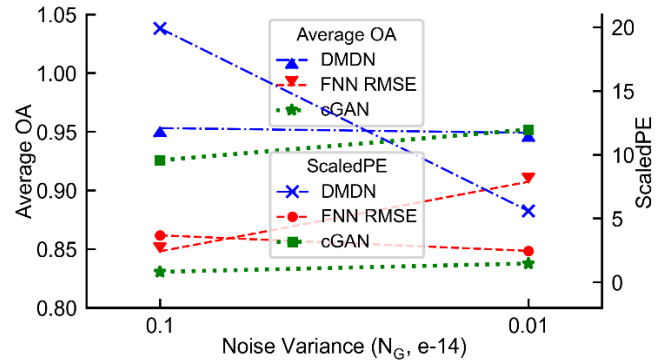

Fig. 3 *Average OA* and *ScaledPE* (left and right y-axes) among FNN with MSE & RMSE and cGAN trained with different data sizes.

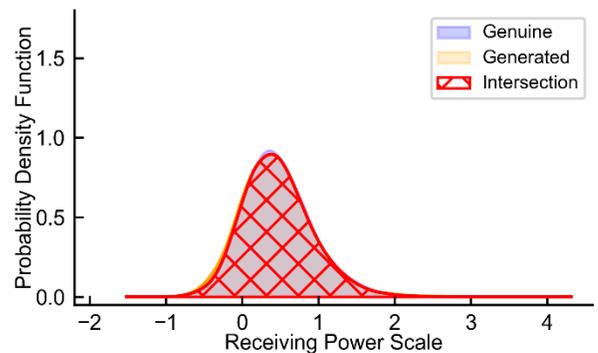

Fig. 4 *OA* of $P_r(d)$'s PDF between the genuine data and the generated data from DMDN for *Nakagami* 1 when $d = 250m$ and $N_{G_{var}} = 1 \times 10^{-15}$

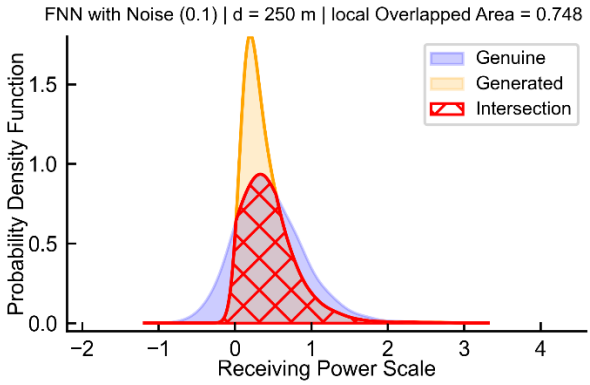

Fig. 5 OA of $P_r(d)'$s PDF between the genuine data and the generated data from FFN for *Nakagami* 1 when $d = 250m$ and $N_{G_{var}} = 1 \times 10^{-15}$

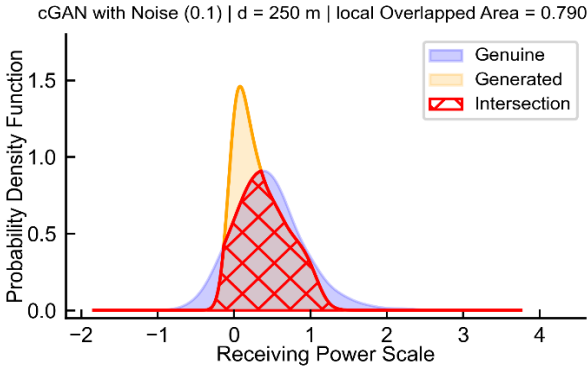

Fig. 6 OA of $P_r(d)'$s PDF between the genuine data and the generated data from cGAN for *Nakagami* 1 when $d = 250m$ and $N_{G_{var}} = 1 \times 10^{-15}$

Fig 4, Fig. 5, and Fig. 6 visualize PDFs of *Nakagami* 1 channel, generated data, and overlapped areas as functions of receiving power scale using three different architecture approaches, respectively. DMDN has the most overlapped area. We can see that DMDN's generated data in Fig. 4 has longer tails than the genuine one on both sides. cGAN in Fig. 6 has a longer tail on the left side than DMDN.

### B. DMDN with Log-Normal Shadowing Channel Model

We apply DMDN to the modeling of the *Log-Normal* 1 (*LN*1) shadowing channel with the following configuration for the urban area traffic environment: $d_0 = 1$ meter, $d_c = 102$, $\delta_1 = 3.9$dB, $\delta_2 = 5.2$dB $\alpha_1 = 2.56$, $\alpha_2 = 6.34$. Other parameters: $P_t = 0.28183815$ Watts (transmission power of each node), $\eta = 7.29 \times 10^{-10}$ (transceiver-determined path loss constant), and $d_{max} = 300$m. Fig. 7 depicts a PDF of the *Log-Normal* 1 channel, generated data from DMDN, and overlapped area of PDF as functions of receiving power scale with the above parameters. From Fig. 7, we can see that the DMDN approach matches the original *Log-Normal* PDF curves very well.

Fig. 8 shows the *Average MOA* as a function of learning epochs starting from random weights. It is observed that DMDN has relatively exceptional training speed (it takes around 15 epochs, ~150 seconds, to converge). Compared to FNN and cGAN, DMDN can handle larger batch sizes. Consequently, we conclude that DMDN has significant potential for real-time training and adjustments with minimal computational demands.

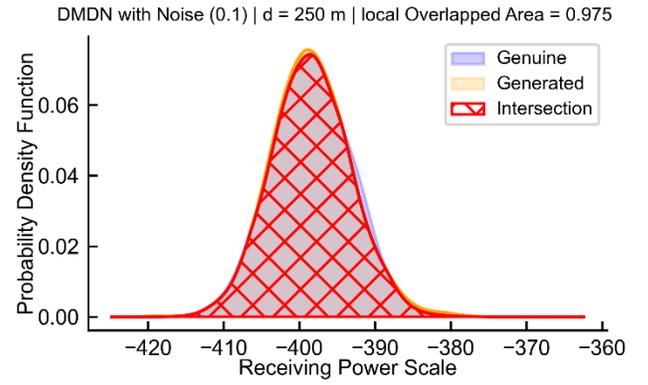

Fig. 7 OA of $P_r(d)'$s PDF between the genuine data and the generated data from DMDN for *Log-Normal* 1 when $d = 250m$ and $N_{G_{var}} = 1 \times 10^{-15}$.

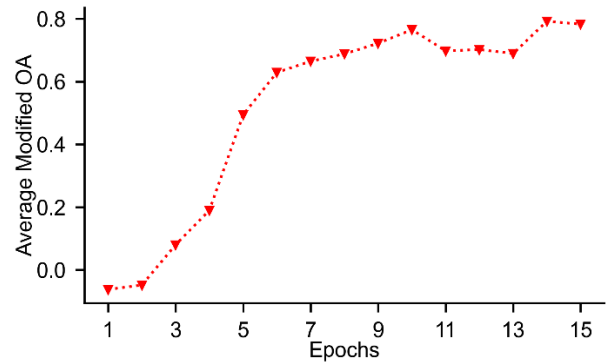

Fig. 8 Modified OA patterns for *Log-Normal* 1 starting from random initialization

### C. Adaptability of DMDN to Dynamic Environments

In this research, in addition to *Nakagami* 1, we trained DMDN to catch up with *Nakagami* 2 (*N*2) channel (same parameters as *Nakagami* 1 (*N*1) except $m = 1.5$ if $d \leq 140$ else $m = 1, \alpha = 2.5$) as well as *LN* 1 for urban communication environments and Log-Normal 2 (*LN* 2 for rural communication environments with parameter setting: $d_0 = 1$meter, $d_c = 182$, $\delta_1 = 3.1$dB, $\delta_2 = 3.6$dB, $\alpha_1 = 1.89$, $\alpha_2 = 5.86$) with and without transfer learning. Transfer learning is designed as follows: 1) Load an existing model (Global Best or Median). 2) Train the model with new data using the same process. Transfer learning in terms of machine learning means training deep neural networks not from scratch (random initialization) but from an existing model (pre-trained) with the same architecture but with different data and tasks. We used two different pre-trained models for transfer learning: Global Best and Global Median. The Global Best model is the model with the best *MOA* values over 15 (epochs) × 10 (iterations) = 150 models on the same training data. Global Median is the model with median ranked *MOA*.

Fig. 9 shows the training convergence curves of *Nakagami* channel modeling regarding the *Average MOA*. It is observed that DMDN reaches the optimal fit remarkably fast when the channel distribution is simple and less complex. Given a DMDN trained with *Nakagami* 1 (*N*1) data, the channel environment is changed to *Nakgami* 2 (*N*2). It takes about three epochs for DMDN to converge to *N*2 if the DMDN is trained with random initial weights. The DMDN instantly converges to the optimal fit if Global Best or Global Median is applied.

Fig. 10 shows training convergence curves of *Log-Normal* channel modeling in terms of the *Average MOA*. DMDN is trained to capture *Log-Normal* 2 (*LN*2) channels starting from DMDN with initial random weights, *Log-Normal* 1 (*LN*1) Global Best, and Global Median. Although the models with Log-Normal data (Fig. 10) have relatively low *Average MOA* compared to those with *Nakagami*, the best models for each training or transfer learning have good *Average OA* and *MOA* values. According to Fig 10, transfer learning from different but less complex or similar channels helps the model to learn new channel distribution effectively.

Table 1 demonstrates the *Average OA* and *MOA* of DMDN channel models with random initialization and different transfer learning schemes. When DMDN reaches an optimal fit, DMDN can hold remarkable performance for both predicting channel distribution for each *d* and on its consistency. According to Table 1, DMDN achieves good results even with random initialization.

## V. CONCLUSIONS

In this paper, we applied a stochastic model, DMDN, to catch different channel distributions. The characteristic of this model aligns well with the nature of the channel model being tested. We discovered that DMDN maintains the consistency of performance regardless of distance and noise and thus overcomes the drawbacks of previously established models (FNN and cGAN). Also, our research implies the potential of DMDN in real-time training and optimization while maintaining its performance and consistency since the model can learn the channel distribution significantly faster with fewer computational resources than the other models. It is concluded that employing transfer learning with DMDN, trained initially on more general or simpler datasets, is recommended for complex channels. Transfer learning may not be necessary for channels with less complex distributions.

Our future research will focus on modeling more general channels and establishing advanced metrics that can objectively measure the model's performance.

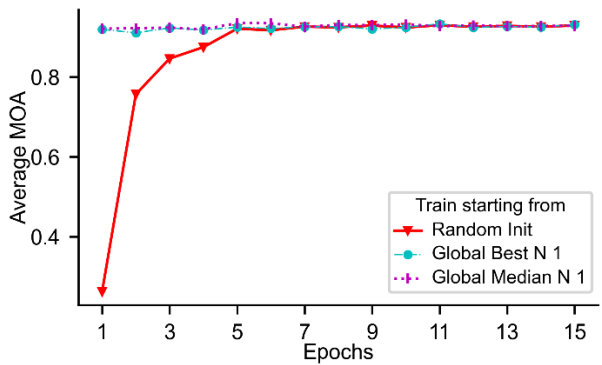

Fig. 9 *Average MOA* DMDN trained with *Nakagami* 2 data starting from random initialization, Global Best, and Median trained with *Nakagami* 1 data.

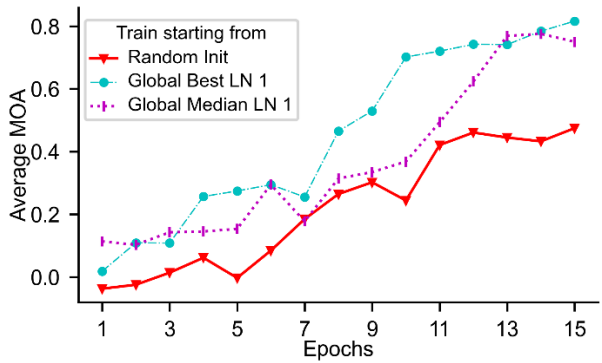

Fig. 10 *Average MOA* DMDN trained with *Log-Normal* 2 data starting from random initialization, Global Best and Median trained with *Log-Normal* 1 data.

Table 1 *Average OA* and *MOA* of DMDN for *Nakagami* 1 & 2 and *Log-Normal* 1 & 2 from the different given statuses.

| Trained with | From | *Average OA* | *MOA* |
|---|---|---|---|
| N1 | Random Init | 0.972 | 0.953 |
| N2 | Random Init | 0.970 | 0.946 |
|  | Global Best N1 | 0.970 | 0.947 |
|  | Global Best N1 | 0.972 | 0.953 |
| LN1 | Random Init | 0.966 | 0.940 |
| LN2 | Random Init | 0.954 | 0.905 |
|  | Global Best LN1 | 0.973 | 0.952 |
|  | Global Best LN1 | 0.957 | 0.925 |